# Training a U-Net based on a random mode-coupling matrix model to recover acoustic interference striations

Xiaolei Li,1, a) Wenhua Song,2 Dazhi Gao,1, a) Wei Gao,1 and Haozhong Wang1

1) *Department of Marine Technology, Ocean University of China, Qingdao, 266100, China*

2) *Department of Physics, Ocean University of China, Qingdao, 266100, China*

*lxl_ouc@outlook.com,*

*songwenhua@ouc.edu.cn,*

*dzgao@ouc.edu.cn,*

*gaowei@ouc.edu.cn,*

*coolice@ouc.edu.cn*

**Abstract:** A U-Net is trained to recover acoustic interference striations (AISs) from distorted ones. A random mode-coupling matrix model is introduced to generate a large number of training data quickly, which are used to train the U-Net. The performance of AIS recovery of the U-Net is tested in range-dependent waveguides with nonlinear internal waves (NLIWs). Although the random mode-coupling matrix model is not an accurate physical model, the test results show that the U-Net successfully recovers AISs under different signal-to-noise ratios (SNRs) and different amplitudes and widths of NLIWs for different shapes.

1. Introduction section

With the development of deep learning (DL), researchers have begun to use DL methods to address underwater acoustic problems, such as source localization [1-6] and target detection [7]. DL can achieve specific functions by fitting training data. Training data are obtained by either experiments or simulations. In previous DL-related work [3-6], only range-independent ocean waveguides were considered during simulations, which is far from the real ocean waveguides in many cases. It is difficult to prepare training data in range-dependent waveguides. On the one hand, it is time-consuming to compute the sound field in a range-dependent ocean waveguide. On the other hand, it is difficult to describe different range-dependent waveguides with finite parameters. For example, even if the range dependence of an ocean waveguide is only caused by a nonlinear internal wave (NLIW), it is difficult to describe all types of NLIWs with finite parameters. Because of the lack of training data, few researchers have used DL to address related problems in range-dependent ocean waveguides.

One of the related problems in range-dependent ocean waveguides is to recover acoustic interference striations (AISs) from distorted ones. AISs, which are related to waveguide-invariant $\beta$ in a range-independent background ocean waveguide, have been used in source ranging [8-11]. If horizontal inhomogeneity is introduced into the background ocean waveguide, for example, range-dependent waveguides with NLIWs, AISs will be distorted due to mode coupling [12,13]. In contrast to the $\beta$ distribution of an AIS, the $\beta$ distribution of a distorted AIS may have several peaks [13], which impedes waveguide-invariant related applications. The purpose of AIS recovery is to eliminate the mode-coupling effect caused by the horizontal inhomogeneity and to recover the corresponding AIS from distorted AIS. Although how an AIS becomes a distorted one by mode coupling has been studied [12,13], AIS recovery still remains unsolved in underwater acoustics. In fact, AIS recovery is to build a mapping from distorted AISs to the corresponding AISs. However, the relationship between AIS and distorted AIS is unknown. With the help of nonlinear activation functions and a large number of neural network parameters, DL is powerful for mining features or relationships from data, which is invaluable in the context of big data, as it extracts high-level information from huge volumes of data. Please refer to Goodfellow et al. [14] for a good textbook of DL. Then AIS recovery can be realized by training a deep neural network (DNN) with large amounts of training data.

In this letter, a U-Net [15], which is a type of DNN, is trained to achieve AIS recovery. To overcome the lack of training data for AIS recovery, a random mode-coupling matrix model is introduced. Training data for AIS recovery can be prepared efficiently by the random mode-coupling matrix model. To verify the effectiveness of the random mode-coupling matrix model, the performance of AIS recovery of the U-Net is tested in range-dependent waveguides with NLIWs. The test results show that the trained U-Net can achieve AIS recovery under different signal-to-noise ratios (SNRs) and different amplitudes and widths of NLIWs for different shapes.

## 2. Random mode-coupling matrix

The mode-coupling matrix method [16,17] is used to describe the mode coupling induced by horizontal inhomogeneities in ocean waveguides. For a receiver located at range $r$ and depth $z$, the pressure field for a narrowband signal can be expressed in terms of normal modes:

$$p(r,z,\omega) = \sum_{m=1}^{M} \frac{A_m(r,\omega)}{\sqrt{k_m(\omega)r}} \varphi_m(z,\omega) \qquad (1)$$

where $\omega$ is the angular frequency of the source, $k_m(\omega)$ and $\varphi_m(z,\omega)$ are the horizontal wavenumber and the mode depth function for the mth mode, $z$ is the depth of the receiver, and $A_m(r,\omega)$ is the (complex) mode amplitude at range $r$. $k_m(\omega)$ and $\varphi_m(z,\omega)$ are computed by Kraken [18] in this letter. The mode amplitude vector

$$\mathbf{A}(r) = [A_1(r,\omega), A_2(r,\omega), \cdots, A_M(r,\omega)]^T,$$

where the subscript denotes the mode number and the superscript ``T'' denotes the matrix transpose, can be expressed as [16,17]

$$\mathbf{A}(r,\omega) = \mathbf{T}(r,r_2,\omega)\mathbf{\Lambda}(r_2,r_1,\omega)\mathbf{T}(r_1,0,\omega)\mathbf{A}(0,\omega), \qquad (2)$$

where $\mathbf{T}(r_1,r_2,\omega) = \mathrm{diag}\{\exp(il_m(\omega)(r_2-r_1))\}$ with $m=1,2,\cdots,M$, is the mode propagation matrix assuming range-independent or adiabatic modes between ranges $r_2$ and $r_1$, and $l_m(\omega) = k_m(\omega) + i\alpha_m(\omega)$, where $\alpha_m(\omega)$ is the mode attenuation of the background ocean waveguide, respectively. $\mathbf{\Lambda}(r_2,r_1,\omega)$ is the mode-coupling matrix which is caused by the horizontal inhomogeneity, such as NLIW, between $r_1$ and $r_2$ [16,17]. Mode-coupling effects introduce many additional components in AISs [13] which results in the distortion of the AISs.

Considering the diversity of horizontal inhomogeneity of ocean waveguide, it is time consuming to calculate the mode-coupling matrix of each type of horizontal inhomogeneity to prepare training data. Note that mode coupling is expressed mainly by the nondiagonal elements of the mode-coupling matrix $\mathbf{\Lambda}(r_2,r_1,\omega)$, and AIS can be recovered by suppressing the influence of the nondiagonal elements. When the region of horizontal inhomogeneity is much smaller than the distance of sound propagation and the variation of ``adiabatic'' mode dispersion caused by the horizontal inhomogeneity can be neglected [19], instead of calculating the mode-coupling matrix of each type of horizontal inhomogeneity, we replace the mode-coupling matrix $\Lambda(r_2, r_1, \omega)$ with a random matrix during training data preparation and assume that mode coupling occurs at a single range, i.e., $r_1 = r_2$ in Eq. (2). In this letter, the diagonal elements of the mode-coupling matrix represent the ``adiabatic'' propagation of each mode. Additionally, considering the narrowband situation and the mode depth functions being weakly dependent on frequency, the dependence of $\Lambda$ on frequency is ignored for simplicity. Referring to Eq. (14) in Ref [16] and assuming that the phase change of diagonal elements in mode-coupling matrix caused by horizontal inhomogeneity can also be neglected, the random mode-coupling matrix $\Lambda$ has the following form:

$$\Lambda_{mn} = a_m \delta_{mn} - i\eta_0 R_{mn}, \tag{3}$$

where $\Lambda_{mn}$ is the element of the nth column in the mth row of $\Lambda$, $a_m$ and $\eta_0$ are positive numbers, $\delta_{mn}$ is the Kronecker $\delta$ function, $R_{mn} \sim N(0,1)$ when $m \neq n$, $R_{mn} = 0$ when $m = n$, and $a_m$ and $\eta_0$ are used to adjust the strength of mode coupling. We introduce the imaginary unit in front of $\eta_0$ in Eq. (3) so that the mathematic form of the random mode-coupling matrix is closer to the first-order expansion form of the mode-coupling matrix [16].

Generally, normal modes of adjacent numbers are more easily coupled with each other, so the coupling matrix often has a band structure [13]. This structure is ignored in the above random mode-coupling matrix model, and the values of non-diagonal elements in $\Lambda$ are independent of each other. Then, Eq. (2) can be rewritten as

$$\mathbf{A}(r,\omega) = \mathbf{T}(r,r_1,\omega)\Lambda\mathbf{T}(r_1,0,\omega)\mathbf{A}(0,\omega), \qquad (4)$$

where mode coupling occurs at $r_1$; see Fig 1 (a).

Using random mode-coupling matrix to model the mode-coupling effect caused by horizontal inhomogeneity, 10000 training samples are generated efficiently. In the process of training the U-Net, distorted AISs are the inputs of the U-Net and the corresponding AISs are the labels. The details of training data preparation, architecture of the U-Net and training process are shown in the supplement material [20]. In the next section, only the results of AIS recovery by the trained U-Net are discussed. The results of AIS recovery by a trained VGG-based convolutional neural network are also shown in the supplement material [20] to indicate that AIS recovery can be achieved by other neural networks too.

## 3. Results of AIS recovery by U-Net

In this section, the U-Net, which recovers AISs from distorted ones, is tested. The schematic diagram of background test environment is shown in Fig 1(a), where $c_b$, $\rho$ and $\alpha$ represent the sound speed, density and attenuation coefficient of the half-space, respectively, $f$ represents frequency in kHz and $\lambda$ is wavelength in m. A horizontal linear array (HLA) that consists of 41 elements with a spacing of 50 m is located at depth of 35 m below the sea surface. A point source with a frequency band of 600-800 Hz is at the same depth as the HLA and $r_S$ from the HLA in the horizontal direction. The sound speed profile is the same as that presented in Fig 5 of Ref [20], which is reproduced here as Fig 1(b). In the test environment, the velocity of the point source is $v_S = 2.4 \text{ m/s}$ and $r_S = (2 \times 10^4 + v_S t) \text{ m}$, where $t \in [0, 2.4 \times 10^4]$ represents time in s. A NLIW, which is created by a depression of the thermocline from 10 m to 19 m, propagates in the direction from the HLA to the source at a velocity of $v_I = 0.6 \text{ m/s}$. The magnitude of $v_I$ coincides with the experimental observation [21]. $r_I = (2 \times 10^3 + v_I t) \text{ m}$. In the propagation of NLIW, the change in the shape of the NLIW is neglected. The sound field received by the HLA is computed by the method proposed by Yang [16] every 10 minutes. Because the velocities of the source and the NLIW are much less than the speed of sound, the movements of the point source and the NLIW are ignored in a single test sample, and the coupling effect between amplitude and speed of the NLIW is not considered.

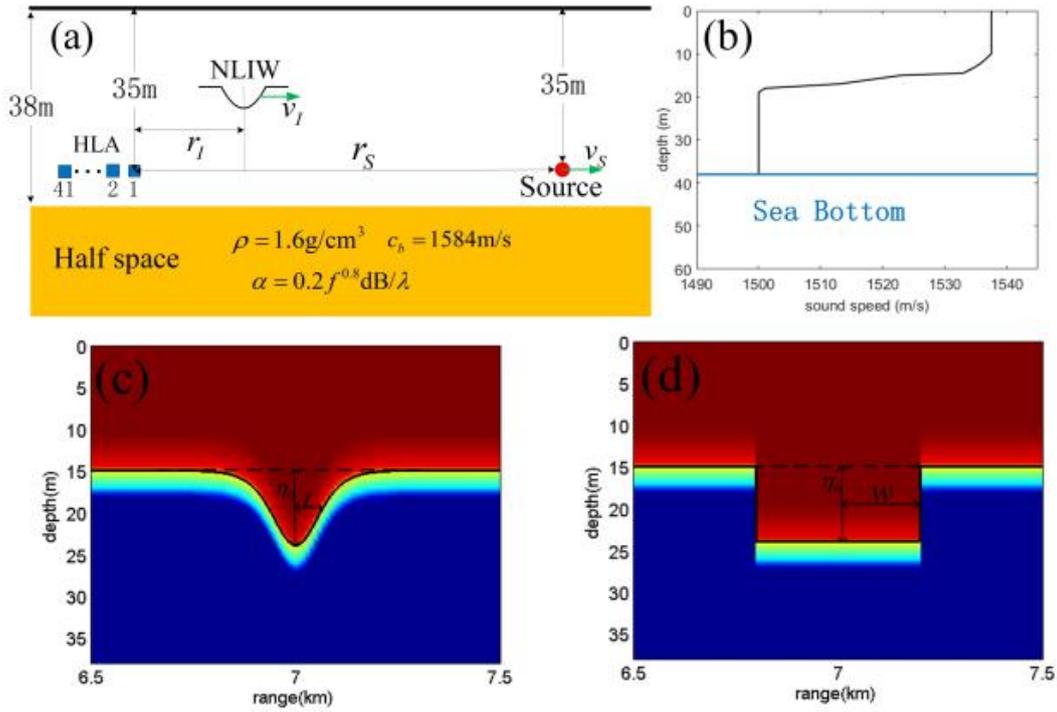

Fig. 1. (color online) (a) Simulation environment schematic diagram. $c_b$, $\rho$ and $\alpha$ represent sound speed, density, and attenuation of the fluid sea bottom. $f$ is frequency in kHz and $\lambda$ is wavelength in meter. A horizontal linear array (HLA) is deployed at a depth of 35 m, which consists of 41 elements with spacing 50 m. A point source with a frequency band of 600-800 Hz is at the same depth as the HLA. The distance between a point source and the HLA is $r_S$. The distance between a nonlinear internal wave (NLIW) and the HLA is $r_I$. (b) Background sound speed profile. (c) Sound speed profile of a Sech–NLIW where $\eta_0 = 9$ m, $L = 75$ m and $r_I = 7$ km. (d) Sound speed profile of a Rect–NLIW where $\eta_0 = 9$ m, $w = 200$ m and $r_I = 7$ km.

Two types of NLIWs are considered in the test: Sech--NLIW and Rect--NLIW. The Sech--NLIW is a NLIW whose shape satisfies

$$\eta(r) = \eta_0 \text{sech}^2\left(\frac{r - r_I}{L}\right) \tag{5}$$

where $\eta_0$ and $L$ determine the amplitude and width of the Sech--NLIW. Rect--NLIW is a NLIW whose shape is rectangular with amplitude $\eta_0$ and half-width $w$. These two types of NLIW are chosen only for analytical convenience, just like James et al. [22] did. During the test, signal to noise ratio (SNR) is 10 dB, $\eta_0 = 9 \text{ m}$, $L = 75 \text{ m}$ and $w = 200 \text{ m}$. SNR is defined by Eq. (3s) in the supplement material [20]. The sound speed profiles of the NLIWs are shown in Fig 1 (c) and (d). The lower-left corner of Fig 2 (a) shows a test sample where $r_S = 35\,\text{km}$, and no striation pattern can be found from the distorted AIS. The lower right corner and the lower center of Fig 2(a) show the corresponding recovered AIS and the corresponding label, respectively. It is found that the recovered AIS is very similar to the label. In Fig 2(b), $E_D(\beta)$, $E_R(\beta)$ and $E(\beta)$ represent the corresponding normalized β distributions of the distorted AIS, the recovered AIS and the label, respectively. Many peaks are observed in $E_D(\beta)$ because of mode coupling, and the distribution is too blurred to determine the β value of the background environment in this case. However, $E_R(\beta) \approx E(\beta)$, and there is only one peak at β =1.5, which indicates that the U-Net has been capable of recovering AIS.

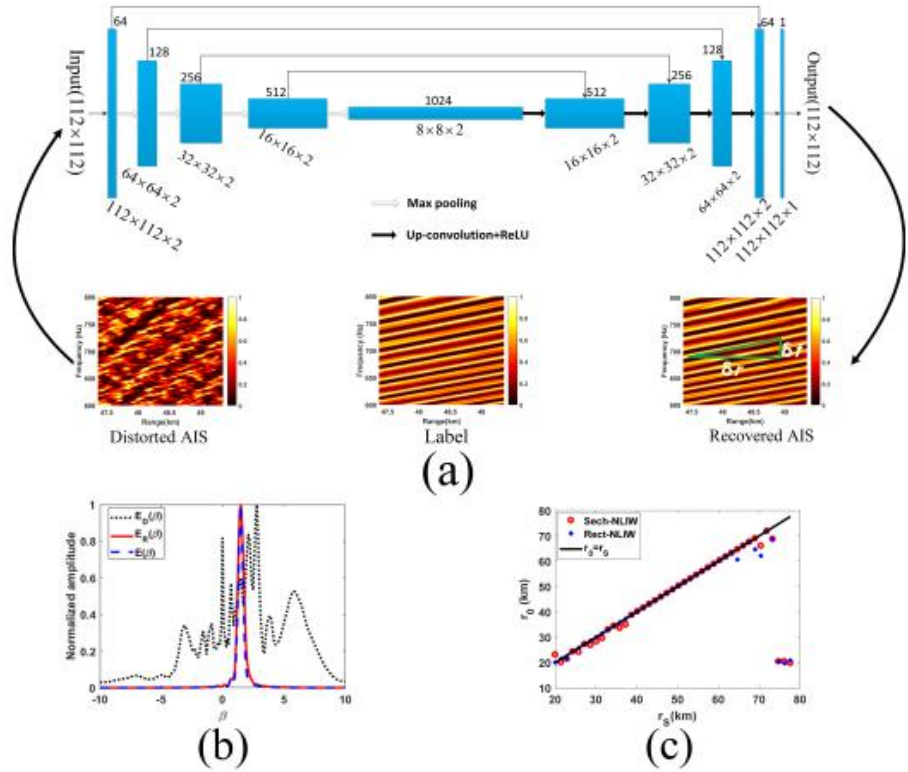

Fig. 2. (color online) (a) A distorted AIS is the input of the U-Net. The output of the U-Net is the recovered AIS. The undistorted AIS is the corresponding label. Each box of the U-Net corresponds to convolution layer(s). The number of channels in each convolution layer is denoted on top of the box. The x-y-size and the number of convolution layers are provided at the lower edge of the box. (b) Normalized $\beta$ distributions of the distorted AIS, the recovered AIS and the label, which are represented by $E_D(\beta)$, $E_R(\beta)$ and $E(\beta)$, respectively. (c) Ranging results by the recovered AIS. Circles are ranging results for the Sech-NLIW case and crosses are ranging results for the Rech-NLIW case.

Given that the waveguide invariant β=1.5, according to the slope of the recovered AIS, one can range the source by the equation below [9]

$$r_0 = \beta \cdot f \cdot \frac{\delta r}{\delta f}, \qquad (6)$$

where $f = 700\,\text{Hz}$ is the central frequency and $\delta r / \delta f$ is the inverse of the slope of the recovered AIS, see Fig 2(a). Fig 2(c) shows that the ranging results are accurate when $r_S \in [20, 73.4]\,\text{km}$. The average relative ranging error, which is defined as $\langle |r_0 - r_S|/r_S \rangle$, is 0.02 and 0.01 for the Sech-NLIW case and the Rect-NLIW case, respectively, where ``$\langle \bullet \rangle$'' represents the sample average. Because in the training data, the distance between sound source and HLA is less than 60 km, however, accurate sound source ranging results can still be obtained in the range of 60-70 km, which indicates that when the distance between sound source and HLA is in the range of 60-70 km, the trained U-Net can still realize AIS recovery.

Table 1. Values of different parameters. The leftmost column indicates the parameters to be tested and their test range. When testing a certain parameter, other parameters are fixed.

| | SNR | $\eta_0$ | L | w |
|---|---|---|---|---|
| $-15\text{dB} \leq \text{SNR} \leq 10\text{dB}$ | | 6m | 75m | 200m |
| $1\text{m} \leq \eta_0 \leq 18\text{m}$ | 10dB | | 75m | 200m |
| $10\text{m} \leq L \leq 150\text{m}$ | 10dB | 6m | | Unrelated |
| $50\text{m} \leq w \leq 750\text{m}$ | 10dB | 6m | Unrelated | |

The U-Net is also tested under different SNRs and different amplitudes and widths of Sech--NLIWs and Rect--NLIWs, respectively when $r_S = 35\,\text{km}$ and $r_I = (2 \times 10^3 + 0.6t)\,\text{m}$, $0 \leq t < 5.5 \times 10^4$ s. During the test, the values of different parameters are shown in Table 1. We use the correlation coefficient of $E_R(\beta)$ and $E(\beta)$ to evaluate the performance of the U-Net, where the correlation coefficient is defined as

$$C_R = \frac{\int_{-10}^{10}[E_R(\beta)-\overline{E}_R(\beta)][E(\beta)-\overline{E}(\beta)]d\beta}{\sqrt{\int_{-10}^{10}[E_R(\beta)-\overline{E}_R(\beta)]^2 d\beta \int_{-10}^{10}[E(\beta)-\overline{E}(\beta)]^2 d\beta}}, \quad (7)$$

$$\overline{E}(\beta) = \frac{1}{20}\int_{-10}^{10} E(\beta)d\beta, \quad (8)$$

and $\overline{E}_R(\beta)$ is calculated by the same method as $\overline{E}(\beta)$. The closer $C_R$ is to 1, the better performance for AIS recovery by the U-Net is achieved. The integral interval of β in Eq. (7) is chosen to be -10 to 10 to cover most of the energy of $E_R(\beta)$, $E(\beta)$ and $E_D(\beta)$.

The first two rows of Fig 3 show $C_D$ and $C_R$ at different SNRs and different amplitudes and widths of Sech-NLIW when Sech-NLIWs are in different positions, where $C_D$ is the correlation coefficient of $E_D(\beta)$ and $E(\beta)$. It is found that except SNR<-3 dB, in all cases, $C_R > 0.93 > C_D$ when the position of the Sech-NLIW does not exceed the region of the horizontal inhomogeneity in training data, i.e. $5\,\text{km} < r_I < 15\,\text{km}$. However, $C_R$ fluctuates from 0 to 1 when $r_I \notin [5,15]\,\text{km}$. When Sech-NLIW is replaced by Rech-NLIW, as shown in the last two rows of Fig 3, similar results can be obtained, which indicates that AIS recovery by the U-Net is insensitive to the shape of the NLIW and proves the effectiveness of random mode-coupling matrix model. As shown in the last row of Fig 3, as the amplitude and width of the Rech-NLIW increase, $C_R$ tends to decrease. This is because with the increasing of amplitude and width of the Rect-NLIW, the variation of ``adiabatic'' mode dispersion caused by the NLIW cannot be neglected, and a detailed discussion is given in section 4 in the supplement material [19].

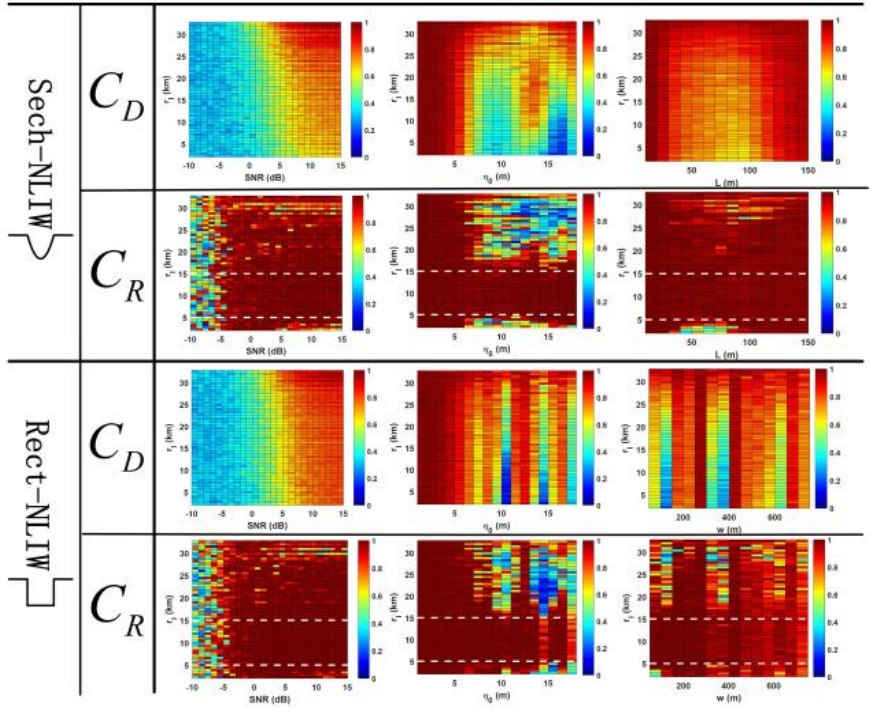

Fig. 3. (color online) $C_D$ and $C_R$ at different SNRs and different amplitudes ($\eta_0$) and widths ($L$ for Sech-NLIW and $w$ for Rech-NLIW) of NLIW when NLIWs are in different positions, where $C_D$ is the correlation coefficients of $E_D(\beta)$ and $E(\beta)$ and $C_R$ is the correlation coefficients of $E_R(\beta)$ and $E(\beta)$. The two dashed lines are $r_I = 5$ km and $r_I = 15$ km, respectively.

## 4. Conclusion

In the letter, a U-Net is trained to realize AIS recovery. To reduce the requiring time in preparing training data, a random mode-coupling matrix model is introduced. Generally, the more kinds of distorted AIS contained in the training data, the better the performance of the trained U-Net. The effectiveness of AIS recovery of the U-Net is verified in range-dependent waveguides with nonlinear internal waves, which indicates that the effectiveness of the random mode-coupling matrix model. Because the variation of ``adiabatic'' mode dispersion caused by horizontal inhomogeneity is neglected in the random mode-coupling matrix model, the performance of AIS recovery is degraded when the variation cannot be neglected. In this work, the uncertainty of background parameters, such as seabed sound velocity and water depth, in the test environment is unconsidered. When there is uncertainty in background environment parameters in the test environment, the uncertainty of the parameters needs to be considered in the training data preparation.

**Acknowledgments**

This work is supported by the National Natural Science Foundation of China under Grant No. 11674294. The authors thank Ning Wang for useful suggestions and discussions and also thank Xinyao Zhang and Xiaohui Sun for useful suggestions on writing.

**References and links**

# Supplementary Material for "Training a U-Net based on a random mode-coupling matrix to recover acoustic interference striation"

Xiaolei Li, Wenhua Song, Dazhi Gao, Wei Gao, and Haozhong Wang

## 1. Training data preparation

The training data used in our work are obtained numerically. The schematic diagram of background environment is shown in Fig. 1s (a), where $c_b$, $\rho$ and $\alpha$ represent the sound speed, density and attenuation coefficient of the half-space, respectively, $f$ represents frequency in kHz and $\lambda$ is wavelength in m. A horizontal linear array (HLA) that consists of 41 elements with spacing 50 m is located at depth of 35 m below the sea surface. A point source with a frequency band of 600-800 Hz is at the same depth as HLA and $r_s$ from the HLA in the horizontal direction. During training date or test data generation, frequency is discretized in increments of 1 Hz. The sound speed profile is shown in Fig. 1s (b). In the generation of test data in the letter, the background environmental parameters of the waveguide are consistent with those here, see Fig. 1 (a) and (b) in the letter.

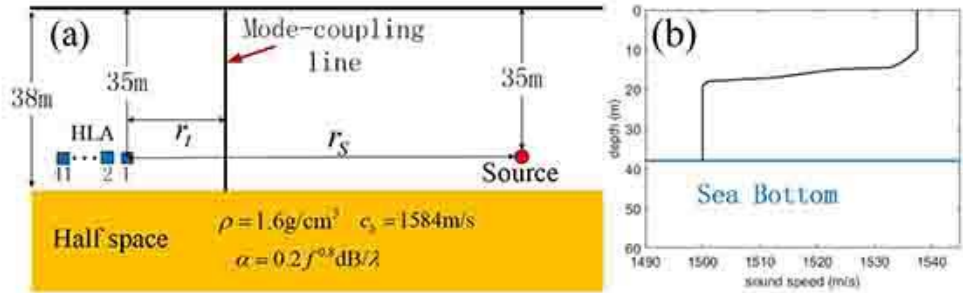

Fig. 1s. (a) Simulation environment schematic diagram. $c_b$, $\rho$ and $\alpha$ represent sound speed, density, and attenuation of the fluid sea bottom. $f$ is frequency in kHz and $\lambda$ is wavelength in meter. A horizontal linear array (HLA) is deployed at a depth of 35 m, which consists of 41 elements with spacing 50 m. A point source with a frequency band of 600-800 Hz is at the same depth as the HLA. The distances between a point source and the HLA and between the position of mode-coupling and the HLA are $r_s$ and $r_l$. (b) Background sound speed profile.

Using $p_A(r_m, f_n)$ and $p_D(r_m, f_n)$ represent the pressure fields at the $m$th element of the HLA in the absence and presence of mode coupling effect in the ocean waveguide, respectively, where $r_m$ is the distance between the source and the $m$th receiver and $f_n$ is the $n$th discrete frequency. $p_A(r_m, f_n)$ is computed by Kraken [1], and $p_D(r_m, f_n)$ is computed by combining Eqs. (1), (3), (4) in the letter, i.e. random mode-coupling matrix method. When $p_D(r_m, f_n)$ is computed, $a_m$ and $\eta_0$ in Eq. (3) satisfy the uniform distributions of 0.5 to 1.5 and of 0 to 5, respectively, to cover the strengths of mode coupling in the test environment.

Then acoustic interference striation (AIS) and distorted AIS, which are represented by $I_A(r_m, f_n)$ and $I_D(r_m, f_n)$ respectively, are calculated, respectively, by Eq. (1s) and (2s).

$$I_A(r_m, f_n) = |p_A(r_m, f_n) / \max(|\mathbf{p}_A(f_n)|)|^2 \tag{1s}$$

and

$$I_D(r_m, f_n) = |[p_D(r_m, f_n) + \sigma n(r_m, f_n)] / \max(|\mathbf{p}_D(f_n) + \sigma \mathbf{n}(f_n)|)|^2, \tag{2s}$$

where $\mathbf{p}_A(f_n) = [p_A(r_1, f_n), p_A(r_2, f_n), \cdots, p_A(r_M, f_n)]^T$ ($\mathbf{p}_D(f_n)$ and $\mathbf{n}(f_n)$ are similar), $M = 41$ is the number of elements of the HLA, $n(r_m, f_n) = A\exp(i\phi)$ represents background noise where $A \sim N(0,1)$ and $\phi \sim U(0, 2\pi)$, and $\sigma$ is a positive number used to adjust the signal to noise ratio (SNR). Both AIS and distorted AIS are normalized at each frequency in Eq. (1s) and (2s) to suppress the influence of the power spectrum of the sound source. In our work, SNR is defined as

$$\mathrm{SNR} = 10\log_{10}\left[\frac{1}{MN\sigma^2}\sum_{m=1}^{M}\sum_{n=1}^{N}|p_D(r_m, f_n)|^2\right], \tag{3s}$$

where $N = 201$ is the number of frequency points. Training data are generated at the SNR of 10 dB. Then, each pair of $r_l$ and $r_s$ corresponds to a training sample, where $r_l$ represents the distance between the mode-coupling line (the line where mode coupling occurs, see Fig. 1s(a)) and the HLA, and $r_s$ represents the distance between the point source and the HLA. The training data set consisting of 10000 training samples is produced by uniformly and randomly selecting $(r_s, r_l)$ in $(20\,\mathrm{km}, 60\,\mathrm{km}) \otimes (5\,\mathrm{km}, 15\,\mathrm{km})$, where "$\otimes$" represents the Cartesian product. A random mode-coupling matrix $\Lambda$ is regenerated for each training sample. After $I_A(r_m, f_n)$ and $I_D(r_m, f_n)$ are computed, they are interpolated into $L \times L$ dimensions to satisfy the input requirement of the corresponding deep neural networks that will be used in the below where $L$ is a positive integer.

## 2. Architecture of U-Net and training process for AIS recovery

We choose a U-Net, which has been widely used in image segmentation [2], to achieve AIS recovery. The architecture of the U-Net is illustrated in the upper panel of Fig. 2s. The U-Net consists of a contracting path (left side) and an expansive path (right side). The contracting path follows the typical architecture of a convolutional network and consists of the repeated application of two 3×3 convolutions, each followed by a rectified linear unit (ReLU) and a 2×2 max pooling operation with stride 2 for downsampling. In each downsampling step, we double the number of feature channels. Every step in the expansive path consists of an upsampling of the feature map followed by a 2 × 2 convolution (up-convolution) that halves the number of feature channels and is followed by a ReLU, a concatenation with the correspondingly cropped feature map from the contracting path, and two 3 × 3 convolutions, each followed by a ReLU. In the final layer, a 1 × 1 convolution followed by a sigmoid function is used to map each 64-component feature vector to the desired number of classes. In total, the network has 23 convolutional layers. $L = 112$ when the U-Net is trained to achieve AIS recovery.

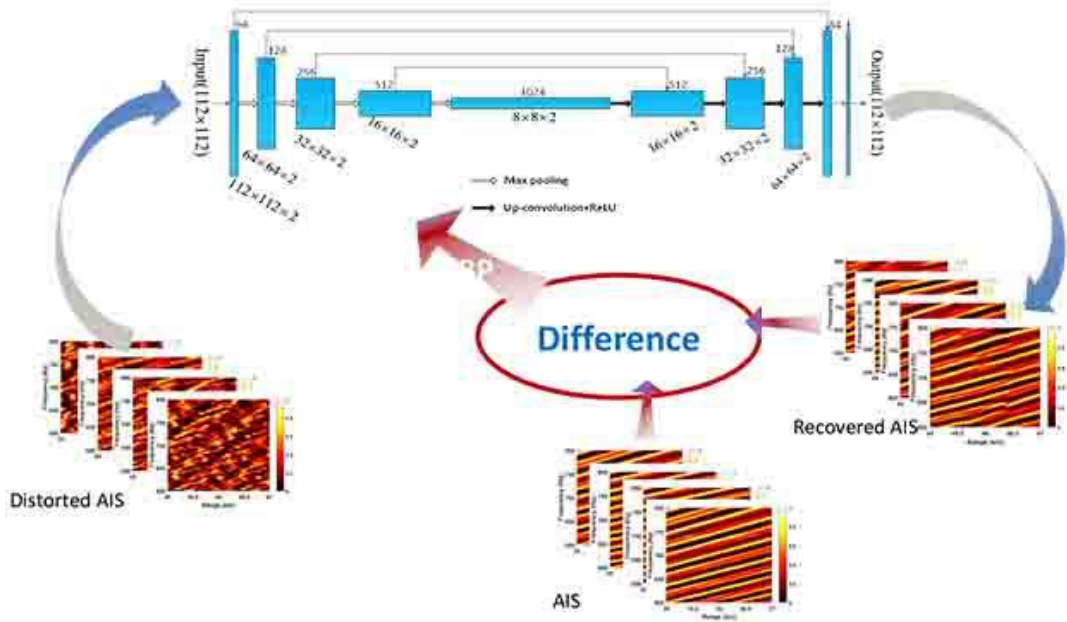

Fig. 2s. Schematic diagram of training a U-Net for AIS recovery. The inputs of the U-Net are distorted AIS, the labels are the corresponding AISs and the outputs of the U-Net are the recovered AISs. "BP" represents the back-propagation method.

When the architecture of a neural network (NN) is fixed, such as a U-Net, the NN can be described as $y = f(x, \theta)$, where x represents the input of the NN and $\theta$ represents the adjustable parameter set of the NN. Using $\hat{y}$ represent the label of x, the training process of the NN regards as the process of minimizing loss function $L(y, \hat{y})$ by adjusting parameter $\theta$ where $L(y, \hat{y})$ is used to describe the difference between y and $\hat{y}$, see Fig.2s. In our work, binary cross-entropy is chosen as $L(y, \hat{y})$.

Combining with back-propagation (BP) method [3], the adaptive moment estimation implementation of Keras [4] with a batch size of 64 is chosen to optimize $\theta$. After 1000 training epochs, the loss function does not decrease and is approximately 0.466, so training is terminated. The results of AIS recovery by the well-trained U-Net are shown in the letter.

## 3. AIS recovery by a VGG-based Convolutional Neural Network

Table 1: Architecture of the VGG-based CNN

| Layer (type) | Output shape | Kernel size | Stride | Activation function |
|---|---|---|---|---|
| Input | (224, 224, 1) | | | |
| Conv2D_1,2 | (224, 224, 64) | (3, 3) | (1, 1) | ReLu |
| MaxPooling2D | (112, 112, 64) | (2, 2) | (2, 2) | |
| Conv2D_3,4 | (112, 112, 128) | (3, 3) | (1, 1) | ReLu |
| MaxPooling2D | (56, 56, 128) | (2, 2) | (1, 1) | |
| Conv2D_5-8 | (56, 56, 256) | (3, 3) | (1, 1) | ReLu |
| MaxPooling2D | (28, 28, 256) | (2, 2) | (2, 2) | |
| Conv2D_9-12 | (28, 28, 512) | (3, 3) | (1, 1) | ReLu |
| MaxPooling2D | (14, 14, 512) | (2, 2) | (2, 2) | |
| Conv2D_13-16 | (14, 14, 512) | (3, 3) | (1, 1) | ReLu |
| MaxPooling2D | (7, 7, 512) | (2, 2) | (2, 2) | |
| DeConv2D_1 | (14, 14, 512) | (2, 2) | (2, 2) | ReLu |
| DeConv2D_2 | (28, 28, 512) | (2, 2) | (2, 2) | ReLu |
| DeConv2D_3 | (56, 56, 256) | (2, 2) | (2, 2) | ReLu |
| DeConv2D_4 | (112, 112, 128) | (2, 2) | (2, 2) | ReLu |
| DeConv2D_5 | (224, 224, 64) | (2, 2) | (2, 2) | ReLu |
| DeConv2D_6 | (224, 224, 1) | (1, 1) | (1, 1) | Sigmoid |

As indicated in the letter, AIS recovery can be achieved by training a neural network (NN). In this section, a VGG-based Convolutional Neural Network (CNN), whose architecture refers to the VGG net [5], is trained to realized AIS recovery, and $I_A(r_m, f_n)$ and $I_D(r_m, f_n)$ are interpolated into $224 \times 224$ ($L = 224$) dimensions to satisfy the input requirement of the VGG-based CNN. The architecture of the VGG-based CNN is shown in Table 1. The distorted AISs and their corresponding AISs are used to train the VGG-based CNN with the adaptive moment estimation implementation of Keras [4] with a batch size of 64, and binary cross-entropy is chosen as the loss function. After 1000 training epochs, the loss function does not decrease and is approximately 0.467, so training is terminated. In the VGG-based CNN, the outputs of one layer only propagate to the next layer, which is the main difference from the U-Net.

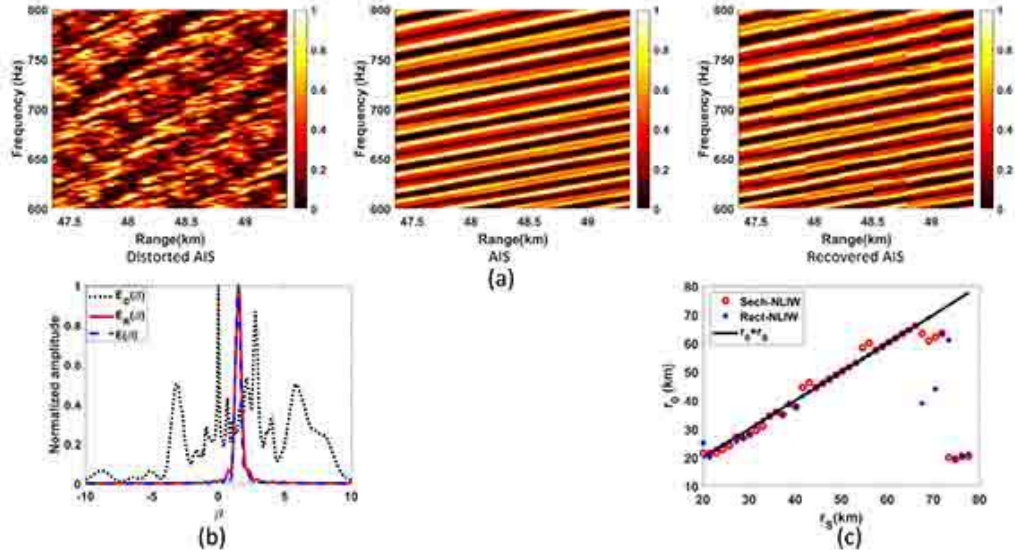

Fig. 3s. (a) A distorted AIS, the corresponding AIS and the recovered AIS by the VGG-based CNN when $r_s = 35$ km. (b) Normalized β distributions of the distorted AIS, the recovered AIS and the label, which are represented by $E_D(\beta)$, $E_R(\beta)$, and $E(\beta)$, respectively. (c) Ranging results by the recovered AIS. Circles are ranging results for the Sech-NLIW case and crosses are ranging results for the Rech-NLIW case.

The test process is the same as in the letter except that the U-Net is replaced by the VGG-based CNN. Fig. 3s(a) shows a test sample when $r_s = 35$ km and no striation pattern can be found from the distorted AIS. Fig. 3s(a) also shows the corresponding AIS and the recovered one, and just like in the letter, one finds that the recovered AIS is very similar to the AIS. In Fig. 3s(b), $E_D(\beta)$, $E_R(\beta)$, and $E(\beta)$ represent the corresponding normalized β distributions of the distorted AIS, the recovered AIS and the label. Many peaks are observed in $E_D(\beta)$ because of mode coupling, and the distribution is too blurred to determine the β value of the background environment in this case. However, $E_R(\beta)$ is similar to $E(\beta)$ and there is only one peak at β = 1.5, which indicates the VGG-based CNN has been capable of recovering AIS. Given that the waveguide invariant β = 1.5, according to the slope of the recovered AIS, one can range the source by Eq. (9) in the letter. Fig. 3s(c) shows that ranging results are accurate when $r_s \in [20,67]$ km with average relative ranging errors are 0.029 for Sech- NLIW case and 0.031 for Rect-NLIW case. Please refer to the letter for the definitions of Sech-NLIW and Rect-NLIW.

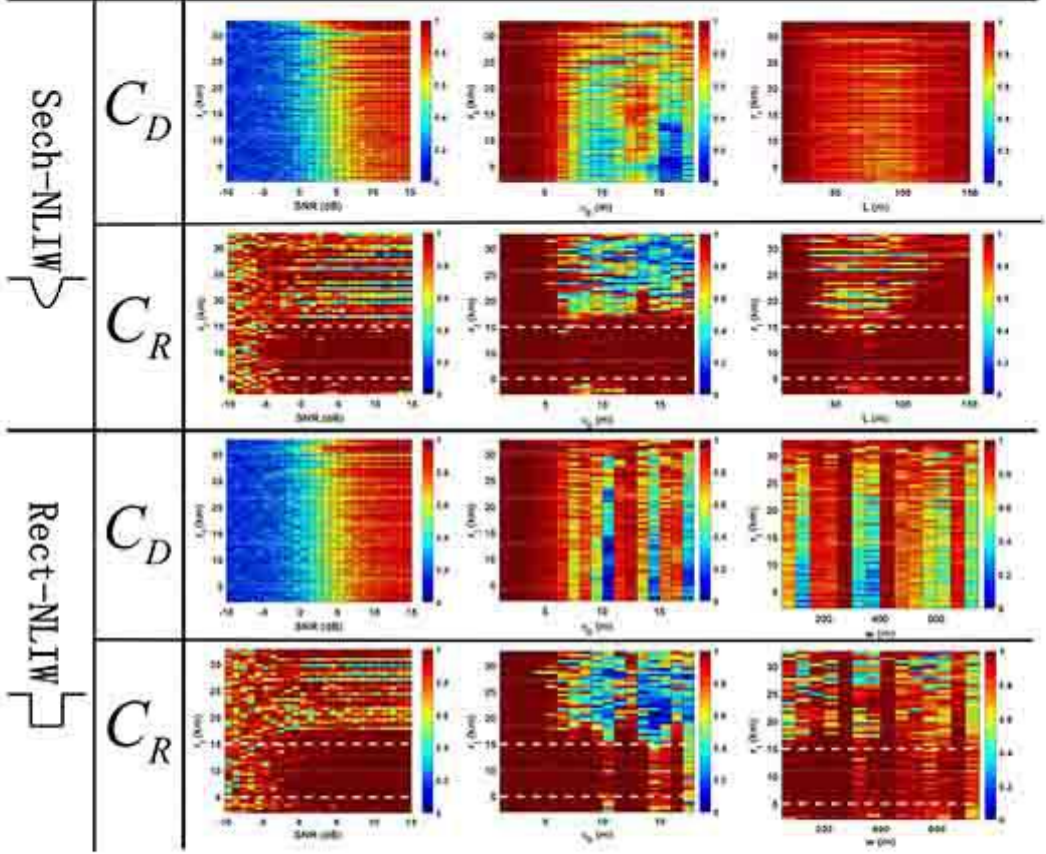

Fig. 4s. $C_D$ and $C_R$ at different signal to noise ratios (SNRs) and different amplitudes and widths of NLIWs when NLIWs are in different positions, where $C_D$ is the correlation coefficient of $E_D(\beta)$ and $E(\beta)$ and $C_R$ is the correlation coefficient of $E_R(\beta)$ and $E(\beta)$. The dashed lines are at $r_I = 5$ km and $r_I = 15$ km, respectively.

Like Fig. 3 in the letter, Fig. 4s shows $C_D$ and $C_R$ at different signal to noise ratios (SNRs) and different amplitudes and widths of NLIWs when NLIWs are in different positions. One can find that the $C_R$ in Fig. 4s is similar to that in Fig. 3 in the letter when $r_I \in [5,15]$ km.

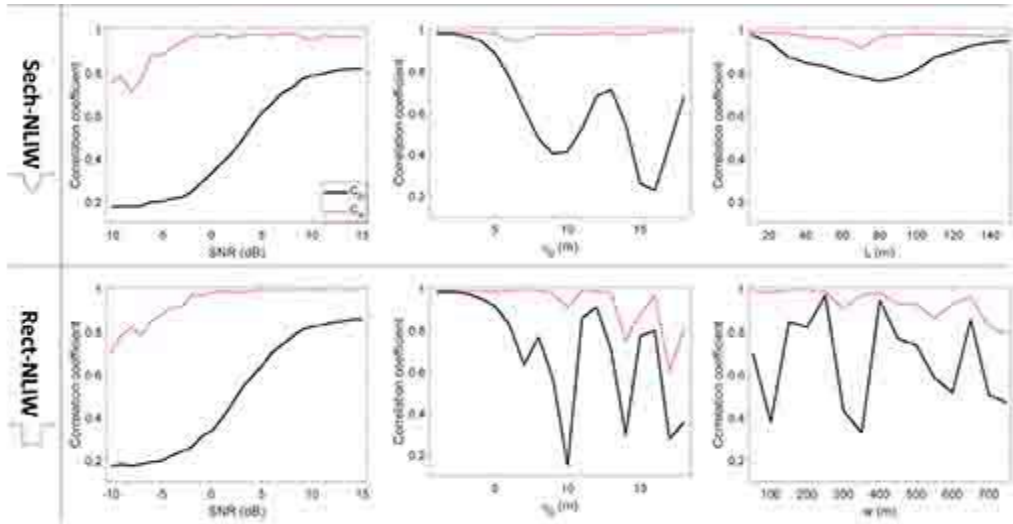

Fig. 5s. When $r_t \in (5,15)$ km, average results of $C_D^*$ and $C_R^*$ about $r_t$ at different SNRs and different amplitudes and widths of NLIWs.

Fig. 5s gives the average results of $C_D^*$ and $C_R^*$ about $r_t$ at different SNRs and different amplitudes and widths of NLIWs when $r_t \in (5,15)$ km. The average value of $C_R^*$ is bigger than that of $C_D^*$. For the Sech-NLIW case, the average value of $C_R^*$ is bigger than 0.9 except when SNR<-4 dB. For the Rech-NLIW case, the average value of $C_R^*$ is bigger than 0.9 except when SNR<-4 dB, the amplitude of NLIW $\eta_0 > 13$ m and the width of NLIW $w$>300 m. When the amplitude of NLIW $\eta_0 > 13$ m and the width of NLIW $w$>300 m, the average of $C_R^*$ has a clear downward trend of oscillation for the Rect-NLIW case. We will explain this downward trend in the next section.

## 4. Explanation of the degradation of AIS recovery performance

In this section, taking Rect-NLIW case as an example, we explain the degradation of AIS recovery performance by the U-Net and the VGG-based CNN. From Fig. 3 in the letter, Fig. 4s and Fig. 5s, one can find that when the amplitude of NLIW $\eta_0 > 13$ m and the half-width of NLIW $w$>300 m, the average of $C_R^*$ has a clear downward trend of oscillation for the Rect-NLIW case. Referring to the energy of the first normal mode, Table 1 gives the relative energy of the first five normal modes received by the first element in HLA where $r_s = 35$ km, f = 700Hz and there is no NLIW. From Table 1, one finds that the AIS is mainly the interference result of the first three normal modes when $r_s = 35$ km. Note that the structure of AIS is related to the dispersion of different modes. The performance of AIS recovery by the U-Net and the VGG-based CNN will degrade when the variation of "adiabatic" mode dispersion caused by Rect-NLIW cannot be neglected, i.e. the Eq. (4s) cannot be satisfied.

Table 1: the relative energy of the first five normal modes received by the first element in HLA when $r_s = 35$ km, f = 700Hz and there is no NLIW.

| Normal mode | 1 | 2 | 3 | 4 | 5 |
|---|---|---|---|---|---|
| Relative energy | 1.0000 | 2.0888 | 0.4325 | 0.0195 | 0.0003 |

$$\arg\{\text{diag}[\mathbf{T}(r_1,r_2,\omega)]\} \approx \arg\{\text{diag}[\Lambda(r_1,r_2,\omega)]\} + \Phi + 2n\pi, \quad (4s)$$

where $\Phi = \text{diag}(\varphi_1, \varphi_2, \cdots, \varphi_M)$ with $\varphi_m$ is a real number independent of $\omega$, $m = 1, 2, \cdots, M$, $m = 0, \pm 1, \pm 2, \cdots$ and the definitions of $\mathbf{T}(r_1, r_2, \omega)$ and $\Lambda(r_1, r_2, \omega)$ are below the Eq. (2) in the letter. Using $\Lambda_{mm}(r_1, r_2, \omega)$ represent the mth diagonal element of the mode-coupling matrix of the Rect-NLIW, then

$$\theta_m(f) = \arg[\Lambda_{mm}(r_1, r_2, \omega)],$$

where $2\pi f = \omega$. Similarly, $\theta_m^0(f) = \arg[T_{mm}(r_1, r_2, \omega)]$, where

$$T_{mm}(r_1, r_2, \omega) = \exp(ik_m |r_1 - r_2|) = \exp(2ik_m w)$$

with $k_m$ being the mth mode wavenumber and $2w = |r_1 - r_2|$. According to Eq. (4s) in the letter,

$$\Delta\theta_m(f) = |\tilde{\theta}_m(f) - \tilde{\theta}_m^0(f)| \approx 2n\pi, \quad n = 0, \pm 1, \pm 2, \cdots,$$

where $\tilde{\theta}_m(f) = \theta_m(f) - \theta_m(f = 600\text{Hz})$ and $\tilde{\theta}_m^0(f) = \theta_m^0(f) - \theta_m^0(f = 600\text{Hz})$.

Fig.6s (a)-(c) show $\Delta\theta_m(f)$ where m=1, 2 and 3, respectively, when the amplitude of Rect-NLIW $\eta_0$ changes. One can find that when $\eta_0 > 13$ m, both $\Delta\theta_1(f)$ and $\Delta\theta_3(f)$ have great changes. This indicates that when $\eta_0 > 13$ m, the performance of AIS recovery by the U-Net will degenerate, which is consistent with our simulation results. $\Delta\theta_2(f)$ has great changes when $\eta_0 = 10\,\text{m}$, which should be responsible for the degradation of the performance of AIS recovery at $\eta_0 = 10\,\text{m}$, see the last row in Fig. 4s. Fig.6s (d)-(e) show $\Delta\theta_m(f)$ where m=1, 2 and 3, respectively, when the half-width of Rect-NLIW $w$ changes. One can find that both $\Delta\theta_2(f)$ and $\Delta\theta_3(f)$ have great changes when $w > 300\,\text{m}$. This indicates that when $w > 300\,\text{m}$, the performance of AIS recovery by the U-Net will degenerate, which is consistent with our simulation results. Although $\Delta\theta_3(f)$ also have great changes at $w = 200\,\text{m}$, the energy of the third normal mode at the receiving position is four times less than that of the second normal mode, so the influence of the phase change on the final result is not significant.

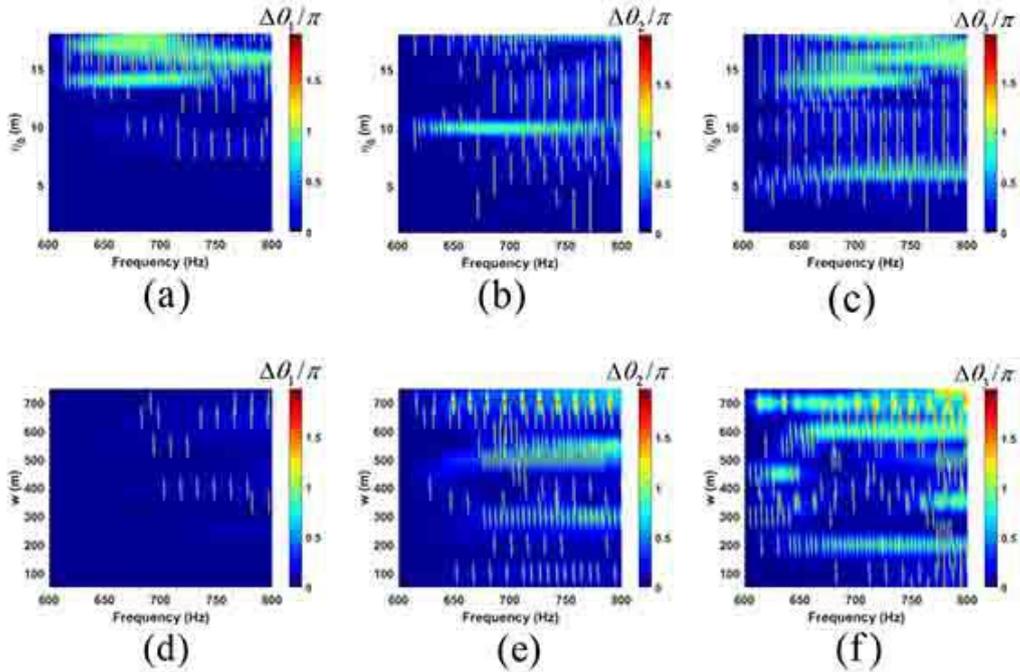

Fig. 6s. Relative phase changes $\Delta\theta_m(f)$ at different $\eta_0$ for m=1, m=2 and m=3 are shown in (a)-(c), respectively. Relative phase changes $\Delta\theta_m(f)$ at different $w$ for m=1, m=2 and m=3 are shown in (d)-(e), respectively.